\documentclass[conference]{IEEEtran}

\usepackage{cite}
\usepackage{amsmath,amssymb}
\usepackage{amsthm}
\usepackage{graphicx}
\usepackage{booktabs}
\usepackage{multirow}
\usepackage{array}
\usepackage{xcolor}
\usepackage{url}

\theoremstyle{plain}
\newtheorem{proposition}{Proposition}
\newtheorem{theorem}{Theorem}
\newtheorem{corollary}{Corollary}
\theoremstyle{definition}
\newtheorem{definition}{Definition}
\newtheorem{assumption}{Assumption}
\theoremstyle{remark}

\begin{document}

\title{PRELUDE: Separating Common Components from Target-Specific Dependencies for Multivariate Time Series Forecasting}

\author{\IEEEauthorblockN{Dongqing Song\IEEEauthorrefmark{1}, Jianyu Xu\IEEEauthorrefmark{2}, Haobin Ding\IEEEauthorrefmark{2}, Xinyi Guo\IEEEauthorrefmark{2}, Jiahao Du\IEEEauthorrefmark{2},\\
Ruiqi He\IEEEauthorrefmark{2}, Ruiqi Liu\IEEEauthorrefmark{1}, Yuntong Xie\IEEEauthorrefmark{3}, and Chao He\IEEEauthorrefmark{2}}
\IEEEauthorblockA{\IEEEauthorrefmark{1}School of Artificial Intelligence, South China Normal University, Foshan, China\\
\IEEEauthorrefmark{2}Aberdeen Institute of Data Science and Artificial Intelligence, South China Normal University, Foshan, China\\
\IEEEauthorrefmark{3}Faculty of Humanities and Arts, Macau University of Science and Technology, Macau, China\\
Corresponding author: Chao He (chaohe@m.scnu.edu.cn)}}

\maketitle

\begin{abstract}
Multivariate time series forecasting often improves the prediction of a target by modeling the interactions among variables, and cross-variable modules such as cross-sectional pooling, inter-variable attention, and graph structures are used to construct such interactions. However, by intervening on the cross-variable reading of existing models at inference time, we find that although these models substantially reduce forecasting error through cross-variable interaction, their reading is nearly identical for all targets: the gain appears to come from dependencies among variables but mainly comes from the common component. We term this phenomenon spurious dependency and attribute it to common-component confounding: present in every variable, the common component affects how reading distinguishes sources, so that reading tends to be shared across targets and concentrates on a few similar averages; dependencies thus make up a small share of the read-in content, and a common component composed of several drivers is only partly captured. Theoretical analysis further shows that the common component and dependencies place different requirements on reading and should therefore be handled in different ways. Accordingly, we propose PRELUDE, a separated-reading framework that estimates the common component holistically through cross-sectional pooling and multiple common factors, and reads dependencies for each target via lead--lag relations in the innovation space obtained by prewhitening. Extensive experiments on 16 real-world datasets from four domains show that PRELUDE effectively mitigates spurious dependency and achieves the best overall performance: it attains the best MSE on 23 of the 24 short-horizon and air-quality settings and reduces the average MSE by 6.1\% relative to the strongest baseline.
\end{abstract}

\IEEEpeerreviewmaketitle

\section{Introduction}
\label{sec:intro}

Multivariate time series forecasting is widely used in energy dispatch, traffic management, weather forecasting, and cloud operations~\cite{zhou2021informer,bai2020agcrn,bi2023pangu,lin2025cloud}. Compared with univariate forecasting, modeling the interactions among variables can often improve the prediction of a target variable: congestion on an upstream road segment foreshadows changes in downstream flow, and pollutants from neighboring cities affect local concentrations through transport~\cite{zhao2024lift,wang2024timexer,wang2020pm25gnn}. To exploit such information, existing methods model the dependencies among variables through cross-sectional pooling~\cite{han2024softs,lin2025tqnet}, inter-variable attention~\cite{liu2024itransformer,zhang2026emaformer,qiu2025duet}, graph structures~\cite{wu2020mtgnn,yu2026cgtfra,hu2025timefilter}, or explicit lead--lag estimation~\cite{zhao2024lift}, and have achieved remarkable progress. Despite their diverse forms, these cross-variable modules share an implicit assumption~\cite{liu2024itransformer,qiu2025duet,zhao2024lift}: aggregating other variables with a learnable interaction module in the raw observation space, trained end to end for the forecasting objective, suffices to capture the dependency structure among variables.

However, by intervening on the cross-variable reading at inference time, we uncover a counterintuitive phenomenon: models substantially reduce forecasting error through cross-variable interaction, yet their cross-variable reading is nearly identical for all targets and does not distinguish the sources on which each target depends. That is, gaining from other variables does not necessarily imply learning the dependencies among them; in scenarios where dependencies matter most, such as short-horizon traffic forecasting, there remains considerable room to exploit the advance information that other variables provide, as shown in Sections~\ref{sec:obs} and~\ref{sec:exp}. We refer to this phenomenon as \emph{spurious dependency}, where the cross-variable gain appears to come from dependencies among variables but actually comes from components that are shared by all variables and do not vary with the target.

\begin{figure*}[t]
\centering
\includegraphics[width=0.95\textwidth]{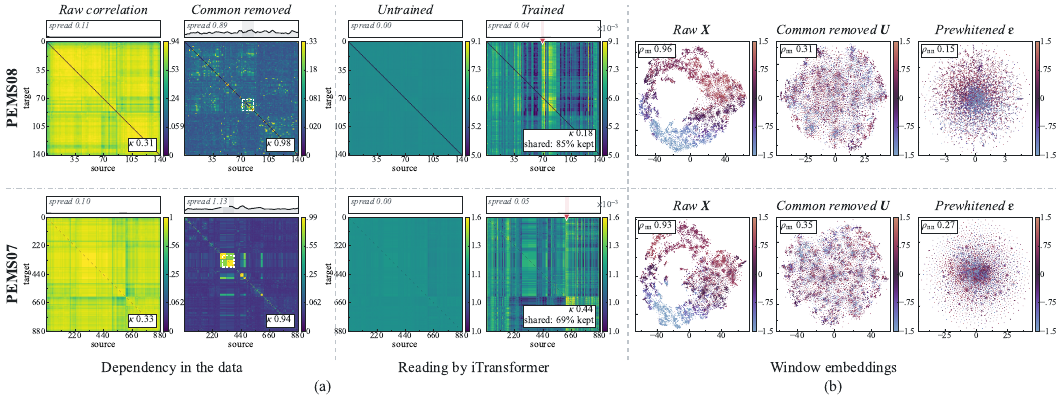}
\caption{Dependencies in the data and the reading of the model (PEMS08 and PEMS07, horizon 3). (a) Relations in the data and the reading of iTransformer (target $\times$ source); the strip above each panel shows the spread of each source's reading weights, and the boxed numbers give the target specificity $\kappa$ and the gain retained under shared reading weights. (b) t-SNE of input windows.}
\label{fig:teaser}
\end{figure*}

To investigate this phenomenon, we conduct a multi-view analysis on PEMS08 and PEMS07 with iTransformer as a representative model (Fig.~\ref{fig:teaser}). \textbf{(1)} After the trend shared by all variables is removed, the dependency in the data is sparse and target specific, with each target related to only a few sources; the trained cross-variable reading, by contrast, concentrates on the same group of sources for all targets and, like an untrained model, does not vary with the target. \textbf{(2)} When all targets share one set of reading weights at inference time, most of the cross-variable gain is retained, whereas it should vanish if the gain came from target-specific dependencies; LIFT, which explicitly estimates lead--lag relations, behaves in the same way. \textbf{(3)} Raw input windows cluster into bands according to the level of the shared trend, and this structure disappears once the trend is removed, indicating that the raw observation space is dominated by the component shared by all variables. Together, these observations show that what existing models obtain from other variables is mainly the trend shared by all variables rather than the dependencies among them. Section~\ref{sec:obs} further shows that this phenomenon is pervasive across seven cross-variable models with different reading forms, and that the shared trend itself is not fully estimated.

Spurious dependency is not accidental. It results from the prevailing practice of learning cross-variable reading in the raw observation space and can be understood from two complementary perspectives. From the data perspective, other variables provide two kinds of information. In a road network, commuting peaks make the flow at all detectors rise and fall together; this part, present in every variable, is called the \emph{common component} and is usually the superposition of several drivers whose proportions differ across variables. An incident at a particular location, in contrast, reaches a few detectors after a certain delay; this part exists only between specific variables and reflects the dependencies among them, whose signal is the \emph{innovation} of the source. Both kinds of information help predict the target, and they are superimposed in the observations with very different magnitudes; since fitting the common component alone already reduces error considerably, spurious dependency often stays hidden behind the improvement in overall error.

From the model perspective, attention, cross-sectional pooling, and graphs all take a weighted average of other variables with row-normalized weights, so a single set of reading weights carries both kinds of information. The common component is present in every source and dominates the scores used to distinguish sources, so reading tends to be shared across targets and concentrates on a few similar averages: the common component enters the read-in content at full amplitude, leaving dependencies only a small share, while the few averages span only a few common directions, so that a common component with heterogeneous proportions is only partly captured. We refer to this mechanism as \emph{common-component confounding}: the common component, as a shared cause of all variables, confounds the apparent relation between source and target. Sections~\ref{sec:obs} and~\ref{sec:theory} confirm this account through interventions and theoretical analysis.

To mitigate spurious dependency and better exploit the information provided by other variables, we propose PRELUDE (PREwhitened Lead--lag Unmasking of DEpendencies), a separated-reading framework for cross-variable forecasting that recasts mixed reading in the raw observation space as holistic estimation of the common component together with dependency reading in the innovation space. The key idea is to estimate the common component holistically over all variables and to read dependencies for each target in the innovation space obtained after removing the common component, so that each kind of information is obtained by its own set of reading weights. PRELUDE first estimates the periodic and multi-factor structure shared by all variables, then removes the common component to prewhiten the observations into innovations, and reads dependencies for each target in the innovation space according to lagged relations, thereby achieving more complete and target-aware cross-variable interaction.

Our contributions are summarized as follows.
\begin{itemize}
\item We identify and characterize spurious dependency in cross-variable forecasting: existing models reduce forecasting error through cross-variable interaction, yet their cross-variable reading does not vary with the target, and the gains come mainly from the common component. We attribute it to common-component confounding and show, through interventions and theoretical analysis, that the common component and dependencies place different requirements on cross-variable reading and should therefore be handled in different ways.
\item We propose PRELUDE, a separated-reading framework that shifts cross-variable modeling from mixed reading in the raw observation space to separated reading, yielding more complete and target-aware cross-variable interaction.
\item We conduct comprehensive experiments on 12 benchmark datasets and on air-quality and short-horizon traffic settings, where PRELUDE achieves the best overall performance; re-measurement with the same diagnostics shows that it estimates the common component more completely and effectively mitigates spurious dependency.
\end{itemize}

\section{Related Work}
\label{sec:related}

\subsection{Cross-Variable Modeling}

According to how they use information across variables, multivariate forecasting methods are commonly divided into channel-independent (CI) and channel-dependent (CD) strategies~\cite{qiu2026survey,qiu2025duet}. CI strategies share one model across all variables and forecast each variable independently: PatchTST splits each variable into patches and encodes them independently~\cite{nie2023patchtst}, and DLinear forecasts the trend and seasonal parts of each variable with linear layers~\cite{zeng2023dlinear}. CD strategies explicitly model the dependencies among variables and can be grouped into four classes according to how they read other variables. Cross-sectional pooling methods aggregate all variables into a shared representation: SOFTS aggregates a core representation through a star-shaped structure and dispatches it back to each variable~\cite{han2024softs}, TQNet reads all variables with learnable periodic queries~\cite{lin2025tqnet}, and TSMixer and TimeMixer mix variables with MLPs~\cite{chen2023tsmixer,wang2024timemixer}. Inter-variable attention methods learn reading weights for each target: Crossformer and iTransformer apply attention along the variable dimension~\cite{zhang2023crossformer,liu2024itransformer}; EMAformer further adds variable identity and phase embeddings~\cite{zhang2026emaformer}; DUET and CCM restrict the reading range of each variable through channel clustering~\cite{qiu2025duet,chen2024ccm}; Crisp blocks interactions between spectrally inconsistent variables with a spectral prior~\cite{wang2026crisp}; and CCD extends variable relations across frequencies and scales~\cite{cheng2026ccd}. Graph-based methods pass information among variables through learned adjacency: StemGNN, MTGNN, CrossGNN, and MSGNet learn graphs among variables~\cite{cao2020stemgnn,wu2020mtgnn,huang2023crossgnn,cai2024msgnet}, CGTFra constrains a dynamically constructed graph with information-theoretic alignment~\cite{yu2026cgtfra}, and TimeFilter filters irrelevant edges on a graph of variables and patches~\cite{hu2025timefilter}. Explicit lead--lag methods estimate dependencies directly: LIFT estimates, for each target, its leading variables and lead steps, and reads them accordingly~\cite{zhao2024lift}. Correlated time series forecasting has also been studied in data engineering, through architecture search, lightweight models, learned graphs, and clustering~\cite{wu2021autocts,wu2023autoctsplus,wu2024autoctsminutes,lai2023lightcts,cirstea2021enhancenet,cirstea2022spatiotemporal,ma2024timeaware,jiang2024sagdfn,niu2025offline,gao2026sparse}, and benchmarked by TFB~\cite{qiu2024tfb}. Although diverse in form, these methods all learn cross-variable reading weights in the raw observation space and train them end to end with the forecasting objective. Building on this line of work, we shift cross-variable modeling from mixed reading in the raw observation space to separated reading, so as to obtain more complete and target-aware interaction.

\subsection{Analysis of Cross-Variable Models}

Several works examine what cross-variable modules actually learn: iTransformer interprets trained attention maps as inter-variable correlations~\cite{liu2024itransformer}, and CPiRi finds that some cross-variable models depend on the ordering of channels~\cite{xu2026cpiri}; Han et al. revisit the trade-off between channel-independent and channel-dependent strategies~\cite{han2024capacity}, and neural Granger causality recovers dependencies from trained models~\cite{tank2022neural}. More broadly, attention weights need not reflect how information actually flows~\cite{abnar2020attention}, and networks tend to exploit strong, simple signals first~\cite{geirhos2020shortcut,shah2020pitfalls,saxe2014exact,pezeshki2021gradient,piao2024fredformer}. This paper starts from where the cross-variable gain comes from: we intervene on cross-variable reading at inference time, measure the gains contributed by the common component and by dependencies separately, and examine our method with the same diagnostics.

\subsection{Modeling the Common Component}

A recent line of work explicitly characterizes the structure shared by all variables. CycleNet models the periodic component of each variable with a learnable recurrent cycle~\cite{lin2024cyclenet}; TQNet introduces this idea into inter-variable interaction through periodic queries~\cite{lin2025tqnet}; TimeEmb characterizes time-invariant and time-varying components with global embeddings and frequency-domain filtering~\cite{xia2025timeemb}; Autoformer, FEDformer, and DLinear decompose series into trend and seasonal parts~\cite{wu2021autoformer,zhou2022fedformer,zeng2023dlinear}; RevIN and Non-stationary Transformers handle shared level shifts through normalization~\cite{kim2022revin,liu2022nonstationary}; and FITS and SparseTSF exploit periodic structure with very few parameters~\cite{xu2024fits,lin2024sparsetsf}. Along this line, we use cross-sectional pooling and multiple common factors to holistically estimate a common component formed by several drivers with different proportions across variables, and prewhiten the observations into innovations with this estimate, so that dependencies can be read in the innovation space.

\section{Preliminaries}
\label{sec:prelim}

\subsection{Problem Statement}

Given a multivariate time series with $N$ variables and look-back length $L$, $X = (X_1, \dots, X_N)^{\top} \in \mathbb{R}^{N \times L}$, multivariate time series forecasting aims to learn a mapping $F_\theta$ with parameters $\theta$ that predicts the next $H$ steps:
\begin{equation}
\hat{Y} = F_\theta(X) \in \mathbb{R}^{N \times H}.
\label{eq:task}
\end{equation}
For a target variable $i$, $X_i \in \mathbb{R}^{L}$ is its look-back window and $Y_i \in \mathbb{R}^{H}$ its future; the remaining variables are called sources.

\subsection{Two Kinds of Information}

\begin{definition}[Common component and innovation]
\label{def:common}
The observation of variable $j$ is written as
\begin{equation}
x_j(t) = c_j(t) + u_j(t), \qquad c_j(t) = p_j(t) + \lambda_j^{\top} f(t),
\label{eq:decomp}
\end{equation}
where $p_j$ is the periodic component, $f(t) \in \mathbb{R}^{r}$ collects $r$ common factors, and $\lambda_j$ is the factor loading of variable $j$; the factors are shared by all variables while the loadings are variable specific, so $c_j$ generally differs across variables; $c_j$ is called the common component and $u_j$ the idiosyncratic component. The part of the idiosyncratic component that cannot be linearly predicted from its own past,
\begin{equation}
\varepsilon_j(t) = u_j(t) - \mathcal{P}\bigl[u_j(t) \mid u_j(t-1), u_j(t-2), \dots\bigr],
\label{eq:innov}
\end{equation}
is called the innovation of variable $j$, where $\mathcal{P}$ denotes the best linear predictor. The transformation from observations to innovations is called prewhitening, and the space in which innovations reside is called the innovation space.
\end{definition}

\begin{definition}[Dependency]
\label{def:dep}
Let $\varepsilon_j^{-}$ denote the innovations of source $j$ up to the current time and $c^{-}$ the common components of all variables up to the current time. The strength of the dependency of target $i$ on source $j$ is defined as the conditional mutual information
\begin{equation}
D_{ij} = I\bigl(Y_i ;\, \varepsilon_j^{-} \mid X_i,\, c^{-}\bigr).
\label{eq:dep}
\end{equation}
If $D_{ij} > 0$, target $i$ is said to depend on source $j$.
\end{definition}

\subsection{Cross-Variable Reading}

\begin{definition}[Cross-variable reading]
\label{def:read}
A cross-variable module gathers information from other variables for target $i$, with the general form
\begin{equation}
m_i = \sum_{j=1}^{N} P_{ij}\, W z_j, \qquad P_{ij} \ge 0, \quad \sum_{j=1}^{N} P_{ij} = 1,
\label{eq:read}
\end{equation}
where $z_j$ is the representation of variable $j$, $W$ is a linear map shared by all variables, $P = (P_{ij})$ are referred to as the reading weights, and $m_i$ is the read-in content of target $i$.
\end{definition}
Attention, cross-sectional pooling, graph modules, and lead--lag estimation are all special cases of this form. When $z_j$ is encoded from raw observations and the same $P$ carries both the common component and the dependencies, we call it \emph{mixed reading}. The reading weights can be decomposed as
\begin{equation}
P = U + \bigl(\mathbf{1}\bar{p}^{\top} - U\bigr) + \bigl(P - \mathbf{1}\bar{p}^{\top}\bigr),
\label{eq:pdecomp}
\end{equation}
where $U$ is the uniform average over other variables ($U_{ij} = 1/(N-1)$ for $j \neq i$) and $\bar{p}_k$ is the mean reading weight of source $k$ across the other targets; the three terms are the uniform average, the deviation shared by all targets, and the target-specific deviation. Over the entries with $i \neq k$, the share of the target-specific deviation in the total deviation,
\begin{equation}
\kappa = \frac{\sum_{k}\sum_{i \neq k}\bigl(P_{ik} - \bar{p}_k\bigr)^2}{\sum_{k}\sum_{i \neq k}\bigl(P_{ik} - \tfrac{1}{N-1}\bigr)^2} \in [0, 1],
\label{eq:kappa}
\end{equation}
is called the target specificity of the reading. Let $E_i(Q)$ denote the forecasting error of target $i$ obtained by substituting $Q$ for $P$ in \eqref{eq:read} at inference time, and let $\mathbf{I}$ denote reading only the target itself. Then
\begin{equation}
\rho_i = \frac{E_i(\mathbf{I}) - E_i(\mathbf{1}\bar{p}^{\top})}{E_i(\mathbf{I}) - E_i(P)}
\label{eq:rho}
\end{equation}
is called the target-invariant share of the cross-variable gain.

\begin{definition}[Spurious dependency]
\label{def:spurious}
A model is said to exhibit spurious dependency for target $i$ if its cross-variable gain $E_i(\mathbf{I}) - E_i(P) > 0$, $\rho_i$ is close to 1, and there exists a source $j$ with $D_{ij} > 0$ in the data.
\end{definition}

\begin{definition}[Common-mode rejection ratio]
\label{def:cmrr}
Apply the same perturbation to the inputs of all variables and denote the resulting change of the read-in content by $\Delta m_i^{\mathrm{c}}$; apply it only to source $k$ and denote the change by $\Delta m_i^{\mathrm{d},k}$. The common-mode rejection ratio (CMRR) of the reading is defined as
\begin{equation}
\mathrm{CMRR}_{ik} = 20 \log_{10} \frac{\lVert \Delta m_i^{\mathrm{d},k} \rVert}{\lVert \Delta m_i^{\mathrm{c}} \rVert}.
\label{eq:cmrr}
\end{equation}
\end{definition}
Row-normalized nonnegative reading has a CMRR of at most 0~dB, and uniform averaging yields $-20\log_{10}(N-1)$~dB.

\subsection{How Much of Each Kind of Information Is Obtained}

\begin{definition}[Residual information spectrum and innovation utilization rate]
\label{def:spectrum}
Let $e_i = Y_i - \hat{Y}_i$ be the residual of the model, let $\mathcal{H}_1$ be the periodic phase of the time stamp, let $\mathcal{H}_2$ add the past of the common factors to $\mathcal{H}_1$, and let $\mathcal{H}_3$ add the past of the source innovations to $\mathcal{H}_2$. The residual information spectrum is defined as
\begin{equation}
s_k = \frac{\mathbb{E}\lVert e_i - \Pi_{k-1} e_i \rVert^2 - \mathbb{E}\lVert e_i - \Pi_k e_i \rVert^2}{\mathbb{E}\lVert e_i \rVert^2}, \qquad k = 1, 2, 3,
\label{eq:spectrum}
\end{equation}
where $\Pi_k$ is the linear projection onto the span of $\mathcal{H}_k$ and $\Pi_0 = 0$; $s_1$, $s_2$, and $s_3$ are called the phase, common-factor, and source-innovation segments. The first two segments correspond to the common component remaining in the residual, and the third to the dependency remaining in the residual. The fraction of the dependency used by the model is called the innovation utilization rate:
\begin{equation}
\mathrm{IUR}_{ij} = \frac{\mathrm{RIG}_{ij}}{\mathrm{RIG}_{ij} + \mathrm{AIG}_{ij}},
\label{eq:iur}
\end{equation}
where the realized gain $\mathrm{RIG}_{ij}$ is the relative increase in the target error when the innovation of source $j$ is replaced at inference time by that of a window at the same time of day whole days apart (same phase and marginal distribution), with its common component, the projection onto the first eight principal directions of the training-segment cross-sections, kept unchanged, and the remaining available gain $\mathrm{AIG}_{ij}$ is the relative reduction in the error of a ridge regression when the non-common part of $X_j$ is added on top of the model's forecast and $X_i$, beyond that from adding its circularly shifted copy; the regression is fitted, regularized, and scored on the training, validation, and test splits, respectively, and without $X_j$ it matches the model's error.
\end{definition}

\section{Observation and Mechanism}
\label{sec:obs}

On seven cross-variable models with different reading forms, this section examines the pervasiveness of spurious dependency (Section~\ref{sec:obs-pervasive}), how much of each kind of information is obtained (Section~\ref{sec:obs-dilution}), and the origin of the phenomenon (Section~\ref{sec:obs-confound}).

\subsection{Diagnostic Setup}
\label{sec:obs-setup}

\textbf{Models and data.} We diagnose seven cross-variable models with different reading forms: SOFTS and TQNet with cross-sectional pooling, iTransformer and EMAformer with inter-variable attention, CGTFra and TimeFilter with graph structures, and LIFT with explicit lead--lag estimation, all trained with their official configurations. The core settings are PEMS08 with horizon 3 and the Yangtze River Delta (YRD) air-quality data with horizon 3\,h. The strongest source of each target is determined from the data: in the space with the common component removed, it is the source with the highest lagged correlation with the target's remaining error that also yields a gain on the validation segment.

\textbf{Synthetic data.} Sources lead the target with fixed delays, a common component with share $e$ is superimposed on all variables, and the future of the target decomposes exactly into the common component, its own dynamics, and source innovations; in a second group, only the amplitude of the common factors is varied.

\textbf{Measurement.} All quantities are computed on the test segment according to the definitions in Section~\ref{sec:prelim}. All interventions only replace inputs or reading weights at inference time; no model is retrained. On synthetic predictors using a known fraction $\lambda$ of the innovation, the IUR reads 0 and 0.98--0.99 at $\lambda = 0$ and 1, is monotone, and reads 0.18--0.24 at $\lambda = 0.1$; it overestimates in between, so near-zero readings are conservative.

\subsection{Pervasiveness of Spurious Dependency}
\label{sec:obs-pervasive}

\begin{figure*}[t]
\centering
\includegraphics[width=\textwidth]{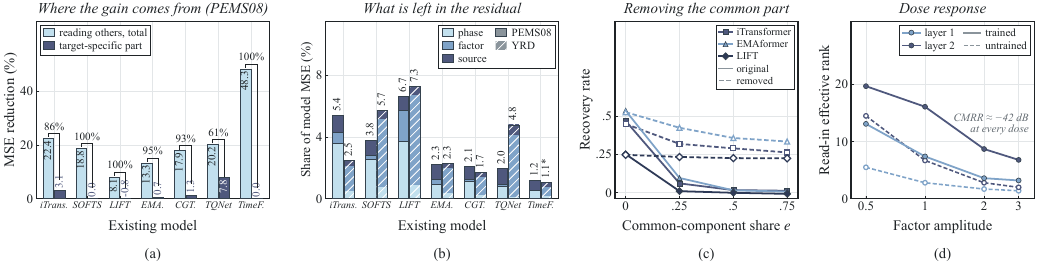}
\caption{What existing models obtain from other variables. (a) Cross-variable gain under progressive replacement of the reading on PEMS08; (b) the part of the residual still predictable from the time of day, the common factors, and a single source; (c) recovery rate of the planted source on synthetic data; (d) effective rank of the read-in content as the common factors strengthen.}
\label{fig:quant}
\end{figure*}

Fig.~\ref{fig:quant}(a) reports the cross-variable gain of the seven models when, at inference time, their reading weights are replaced in turn by the weights shared by all targets, $\mathbf{1}\bar{p}^{\top}$, and by the uniform average $U$. With shared weights, most of the cross-variable gain is retained in all seven models. This holds even for the most concentrated reader: the graph pathway of CGTFra has the highest target specificity among the seven models and selects its own sources for each target, yet retains 94\% of its gain under a uniform average. The concentration and target specificity of reading therefore do not determine where the cross-variable gain comes from. By Definition~\ref{def:spurious}, all seven models exhibit spurious dependency.

\subsection{Diluted Dependencies and Residual Common Components}
\label{sec:obs-dilution}

\begin{figure*}[t]
\centering
\includegraphics[width=0.95\textwidth]{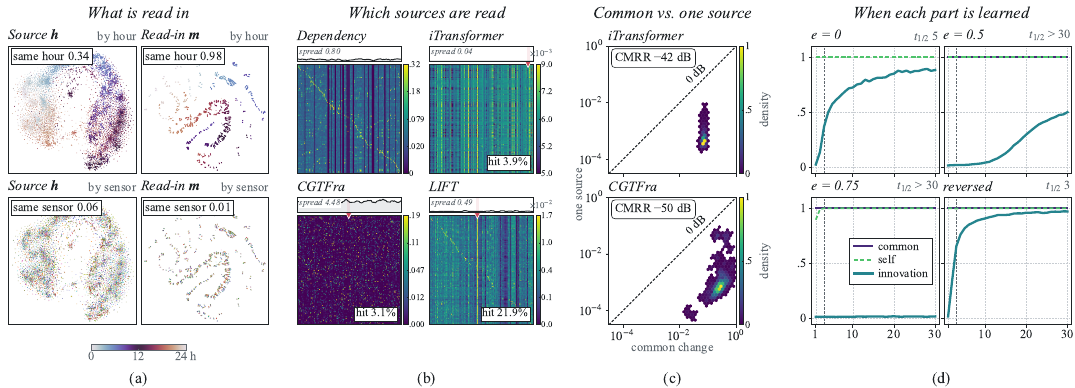}
\caption{Four stages of the reading process in existing models (PEMS08, horizon 3). (a) t-SNE of source embeddings and read-in content; (b) dependency in the data and the reading of three models; (c) response of the reading to a common change and to a single-source change; (d) learning progress of the three components on synthetic data, with the vertical dashed line marking the official training budget.}
\label{fig:reading}
\end{figure*}

\textbf{Diluted dependencies.} When reading does not distinguish sources by target, the common variation of all variables enters the read-in content at full amplitude, whereas the variation of a single source enters only in proportion to its reading weight; the CMRR quantifies the ratio of the two, which is at most 0\,dB for row-normalized reading and $-44.6$\,dB for uniform averaging. As shown in Fig.~\ref{fig:reading}(c), iTransformer responds considerably more strongly to a common-mode perturbation than to a perturbation of the strongest source alone, with a CMRR of $-42$\,dB, close to that of uniform averaging, and the other six models likewise lie below 0\,dB. The read-in content is therefore dominated by the common component, in which the dependency signal makes up a small share, and the innovation utilization rate is close to zero.

\textbf{Residual common components.} How much of the common component mixed reading obtains depends on how its variance is distributed over the common directions. In traffic data the whole network fluctuates synchronously, so the variance of the common component concentrates on a single principal direction and one cross-sectional average can carry most of it; the common component of air-quality data, in contrast, is the superposition of several factors with different loadings, and its variance spreads over several directions. The residual information spectrum in Fig.~\ref{fig:quant}(b) corresponds to this: on traffic data the residual is dominated by the phase segment, whereas on air-quality data it is dominated by the common-factor segment.

Fig.~\ref{fig:reading}(a) corroborates this at the representation level: the share of same-time nearest neighbors rises from 0.34 in the source embeddings of iTransformer to 0.98 in its read-in content, which retains a single common coordinate carrying the principal direction, while the remaining directions stay in the residual.

The dilution of dependencies and the residual common components share a structural feature: reading does not distinguish sources by target and instead concentrates on a few similar averages. Section~\ref{sec:obs-confound} examines the origin of this concentration.

\subsection{Common-Component Confounding}
\label{sec:obs-confound}

Reading distinguishes sources through differences among their scores. The common component is present in every source and enters the scores of all sources through their loadings rather than through dependencies, so it dominates the comparison; innovations, which have small variance and are sparsely distributed, account for only a small part of the differences. It follows that the stronger the common component, the harder it is for reading to identify sources.

Fig.~\ref{fig:reading}(b) is consistent with this inference. In the data, each target depends on only a few sources, and different targets depend on different sources; yet the largest reading weight of iTransformer rarely falls on these sources, with a top-1 hit rate of only 3.9\%. LIFT, which explicitly estimates lead--lag relations, achieves a higher hit rate, but its lagged correlations are computed on raw series; on the raw series of synthetic data, the same statistic recovers only one third of the planted sources, whereas it recovers 96.7\% after the common component is removed.

If the common component is indeed the source of the identification difficulty, removing it should restore the reading of dependencies, and this restoration should occur only when the common component is present. Fig.~\ref{fig:quant}(c) confirms both points: for common-component shares $e \ge 0.25$, removing the common component raises the recovery rate of the planted source by target-wise reading models (the share of the planted gain obtained by the model) two- to fourfold; at $e = 0$, removal makes no difference. This confirms that removing the common component is key to reading dependencies.

Furthermore, when the differences among sources are small, row-normalized reading naturally concentrates on a few similar averages. Fig.~\ref{fig:quant}(d) shows that as the common factors strengthen, the effective rank of the read-in content of iTransformer decreases monotonically, from 13.1 to 3.2 in the first layer; the same architecture at random initialization behaves likewise, so training does not alter this trend. The common component compresses reading into a few directions, which is precisely the shared origin of the two phenomena in Section~\ref{sec:obs-dilution}: a few directions mainly cover the principal direction of the common component, while letting the common component dominate the read-in content at full amplitude.

Nor does this situation change markedly with longer training: Fig.~\ref{fig:reading}(d) shows that the learning of source innovations is delayed as the common component strengthens, with less than half learned after 30 epochs at $e = 0.5$.

In summary, spurious dependency arises from common-component confounding rather than from any particular reading form: whether through cross-sectional pooling, inter-variable attention, graph structures, or lead--lag estimation, when both kinds of information are obtained by one set of reading weights in the raw observation space, the common component dominates the reading and dependencies are diluted. The two kinds of information place different requirements on reading, with the common component calling for consistent, holistic estimation over all variables and dependencies calling for selective, target-wise reading, and should therefore be handled in different ways. This suggests a general principle for cross-variable modeling: estimate the common component as a whole, and read dependencies for each target in the innovation space obtained after removing the common component. Section~\ref{sec:theory} characterizes this principle theoretically, and Section~\ref{sec:method} builds PRELUDE upon it.

\section{Theoretical Analysis}
\label{sec:theory}

Building on the general form of reading \eqref{eq:read} and a factor-structure assumption, this section shows that the chain revealed in Section~\ref{sec:obs} is determined by the structure of mixed reading.

\subsection{Formalization}

\begin{assumption}[Factor structure of representations]
\label{asm:factor}
The representations on which cross-variable reading operates are written as
\begin{equation}
z_j = C\lambda_j + u_j, \qquad j = 1, \dots, N,
\label{eq:factor}
\end{equation}
where $C \in \mathbb{R}^{d \times r}$ is the representation of $r$ common factors, identical for all variables; $\lambda_j \in \mathbb{R}^{r}$ is the loading of variable $j$; and $u_j$ is the representation of the idiosyncratic component, with zero mean and covariance $\sigma_j^2 I$, uncorrelated with $C$. The idiosyncratic components of different variables are uncorrelated except through dependencies: the idiosyncratic component of target $i$ contains the innovation of source $s$ from $\ell$ steps earlier, and the cross-covariance between their representations is $\sigma_I \Gamma_\ell$, where $\sigma_I$ is called the effective strength of the dependency.
\end{assumption}
The attention score is written as $s_{ij} = z_i^{\top} M z_j$ with $M = W_Q^{\top} W_K / \sqrt{d_k}$, and we denote $\Sigma_M = \mathbb{E}[C^{\top} M C]$.

\subsection{The Common Component in the Scores}
\label{sec:theory-score}

\begin{proposition}[Decomposition of score differences]
\label{prop:unident}
Under Assumption~\ref{asm:factor}, for source $s$ and any variable $j$ on which target $i$ does not depend,
\begin{equation}
\mathbb{E}\bigl[s_{is} - s_{ij}\bigr] = \lambda_i^{\top} \Sigma_M (\lambda_s - \lambda_j) + \sigma_I \operatorname{tr}\bigl(M \Gamma_\ell\bigr).
\label{eq:scorediff}
\end{equation}
\end{proposition}
\begin{proof}[Proof sketch]
Substitute \eqref{eq:factor} into $z_i^{\top} M (z_s - z_j)$; the cross terms have zero expectation, and among the idiosyncratic terms only the cross-covariance of $u_i$ and $u_s$ remains.
\end{proof}

The score difference thus consists of a loading term and a dependency term. With identical loadings, the loading term vanishes and sources are distinguished in expectation only by the dependency of strength $\sigma_I$; with different loadings, the loading term depends on the target only through $\lambda_i$, and when it is large relative to the dependency term, sources are ranked by loading similarity rather than by dependency.

\subsection{Two Faces of Mixed Reading}
\label{sec:theory-faces}

\begin{proposition}[Full-amplitude entry of the common component]
\label{prop:fullamp}
Under Assumption~\ref{asm:factor}, for given reading weights $P$, the read-in content is
\begin{equation}
m_i = C\bar{\lambda}_i + \sum_{j} P_{ij} u_j, \qquad \bar{\lambda}_i = \sum_j P_{ij} \lambda_j;
\label{eq:readin}
\end{equation}
with identical loadings, the common component enters $m_i$ with gain 1, regardless of the reading weights. For one-dimensional representations ($d = 1$), let $a_i^2 = \operatorname{Var}(C\bar{\lambda}_i)$; the fraction of the idiosyncratic component $u_s$ of source $s$ that can be linearly read out from $m_i$ is
\begin{equation}
\pi_{is} = \frac{P_{is}^2 \sigma_s^2}{a_i^2 + \sum_j P_{ij}^2 \sigma_j^2}.
\label{eq:pi}
\end{equation}
\end{proposition}
\begin{proof}[Proof sketch]
Substitute \eqref{eq:factor} into \eqref{eq:read} and use $\sum_j P_{ij} = 1$ to obtain \eqref{eq:readin}; $\operatorname{Cov}(m_i, u_s) = P_{is}\sigma_s^2$ then gives \eqref{eq:pi}.
\end{proof}

For source $s$ to be read out, its weight must outweigh both the idiosyncratic components of all other sources and the entire common component; under uniform averaging with equal variances, $\pi_{is} \le 1/(N-1)$, and the 0\,dB bound on the CMRR of row-normalized reading follows directly from \eqref{eq:readin}.

\begin{proposition}[Truncation of common directions]
\label{prop:trunc}
Let the common component be $c_j = \lambda_j^{\top} f$ with $\operatorname{Cov}(f) = \Sigma_f$ and $\Lambda = (\lambda_1, \dots, \lambda_N)^{\top}$, and let $\mu_1 \ge \dots \ge \mu_N$ be the eigenvalues of $\Sigma_c = \Lambda \Sigma_f \Lambda^{\top}$. If reading is target agnostic and the read-in content of all targets is given by $q$ sets of weights, then, when the common component of each target is reconstructed from linear combinations of the read-in content, the ratio of the total reconstructable variance to the total variance of the common component satisfies
\begin{equation}
\frac{\sum_i R_i^2 \operatorname{Var}(c_i)}{\sum_i \operatorname{Var}(c_i)} \le \frac{\sum_{l \le q} \mu_l}{\sum_{l} \mu_l}.
\label{eq:trunc}
\end{equation}
\end{proposition}
\begin{proof}[Proof sketch]
The common parts of the read-in content lie in a subspace of dimension at most $q$; by the extremal property of the trace, the total reconstructable variance does not exceed the sum of the $q$ largest eigenvalues of $\Sigma_c$.
\end{proof}

When the common component is close to a single principal direction, one average suffices to obtain it almost completely; when loadings are dispersed, the tail $\sum_{l > q} \mu_l$ necessarily remains in the error. By Proposition~\ref{prop:unident}, when the loading term dominates, mixed reading varies little with the target, so $q$ is small.

\subsection{Estimation--Identification Trade-off}
\label{sec:theory-tradeoff}

\begin{theorem}[Estimation--identification trade-off]
\label{thm:tradeoff}
In the setting of Proposition~\ref{prop:fullamp}, for any reading weights with $P_{ij} \ge 0$ and $\sum_j P_{ij} = 1$, let $\nu_i = \sum_j P_{ij}^2 \sigma_j^2 / a_i^2$ be the variance ratio of idiosyncratic to common components in the read-in content. Then, for any source $s$,
\begin{equation}
\pi_{is} \le \frac{\nu_i}{1 + \nu_i},
\label{eq:tradeoff}
\end{equation}
with equality if and only if $P_i$ is nonzero only at $s$.
\end{theorem}
\begin{proof}[Proof sketch]
Substitute $P_{is}^2 \sigma_s^2 \le \sum_j P_{ij}^2 \sigma_j^2 = \nu_i a_i^2$ into \eqref{eq:pi}.
\end{proof}

A small $\nu_i$ gives a clean estimate of the common component and requires spread-out weights, whereas $\pi_{is} \to 1$ requires $\nu_i \to \infty$, that is, a read-in content dominated by source $s$. The two kinds of information therefore place opposite requirements on one set of reading weights. Proposition~\ref{prop:fullamp} and Theorem~\ref{thm:tradeoff} are stated for one-dimensional representations; for multidimensional representations, they hold along every direction shared by the common and idiosyncratic components.

\begin{corollary}[Separated reading]
\label{cor:separate}
If the common component and the dependencies are each obtained by their own set of reading weights, with $P^{c}$ acting on the raw representations and spreading over all variables, and $P^{d}$ acting on the innovation representations from which the common component has been removed, then $\nu_i$ can attain its lower bound while concentrating $P^{d}_i$ on $s$ yields $\pi_{is} = 1$.
\end{corollary}
\begin{proof}[Proof sketch]
In the innovation representations $a_i = 0$, and \eqref{eq:pi} equals 1 when $P^{d}_i$ is the unit vector pointing to $s$.
\end{proof}

The estimation--identification trade-off is thus not inherent in cross-variable modeling but arises from one set of weights carrying both kinds of information. When the common component is removed incompletely, the residual common component in the innovation representations makes $\pi_{is}$ smaller than 1 by~\eqref{eq:pi}; the residual is measured in Section~\ref{sec:exp-ablation}.

\subsection{Predictions and Implications}
\label{sec:theory-pred}

These results correspond to the observations of Section~\ref{sec:obs} (Proposition~\ref{prop:unident} to Section~\ref{sec:obs-confound}, Propositions~\ref{prop:fullamp} and~\ref{prop:trunc} to Section~\ref{sec:obs-dilution}), and Theorem~\ref{thm:tradeoff} further predicts that merging the two sets of reading weights brings back both phenomena at once (Section~\ref{sec:exp-ablation}). Accordingly, the common component should be estimated directly as a whole, and dependencies should be read for each target in the innovation space.

\section{Method}
\label{sec:method}

\subsection{Overview}

\begin{figure*}[t]
\centering
\includegraphics[width=0.95\textwidth]{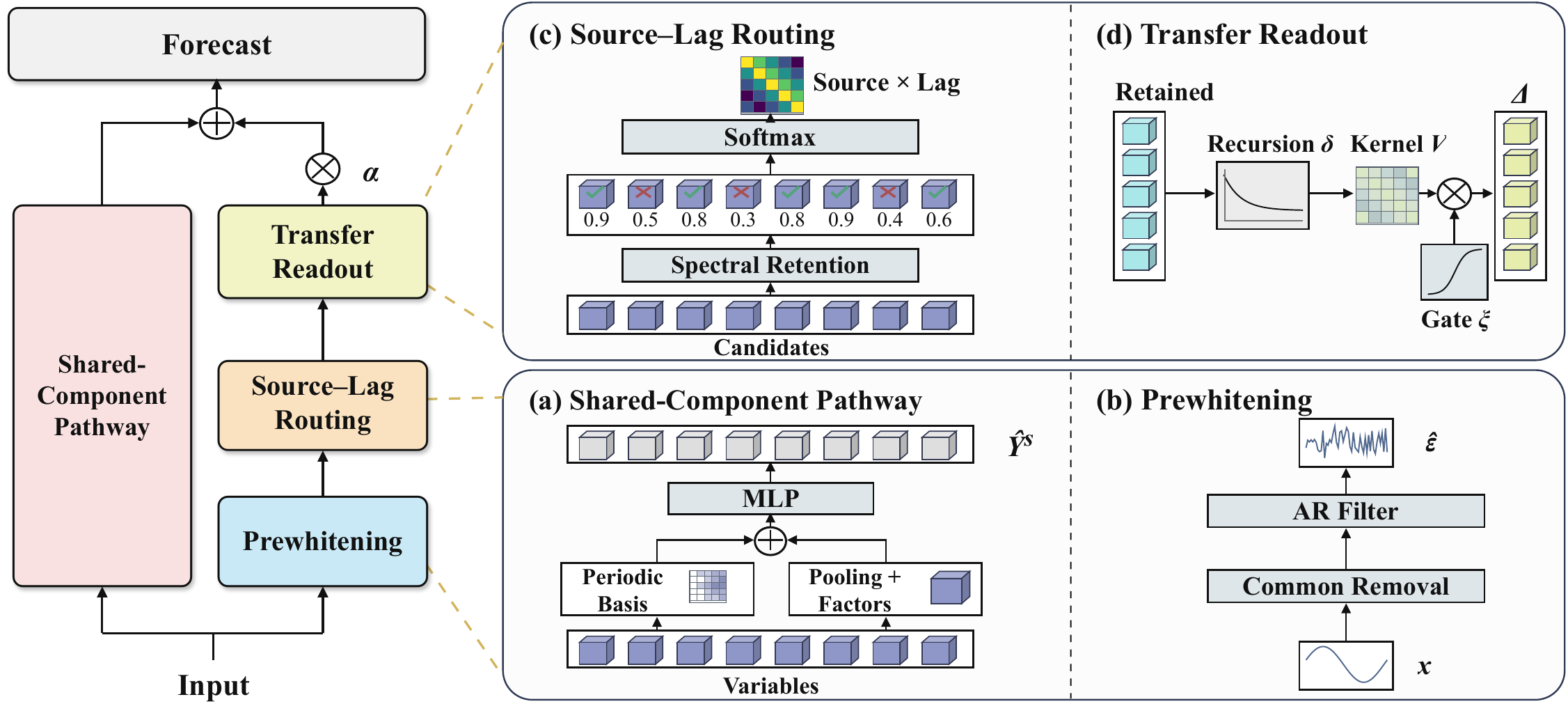}
\caption{Overall architecture of PRELUDE. (a) Shared-component pathway; (b) prewhitening; (c) source--lag routing; (d) transfer readout.}
\label{fig:method}
\end{figure*}

Fig.~\ref{fig:method} shows the overall architecture of PRELUDE. PRELUDE organizes cross-variable modeling as separated reading and consists of three parts: a shared-component pathway (Section~\ref{sec:method-scp}), an innovation pathway (Section~\ref{sec:method-isp}), and a combination weight (Section~\ref{sec:method-alpha}), which respectively perform holistic estimation of the common component, dependency reading in the innovation space, and the combination of the two:
\begin{equation}
\begin{aligned}
\hat{Y}^{s} &= \mathrm{SCP}(X), & \hat{\varepsilon} &= \mathrm{Prewhiten}(X), \\
\Delta &= \mathrm{ISP}(\hat{\varepsilon}), & \hat{Y} &= \hat{Y}^{s} + \alpha \odot \Delta,
\end{aligned}
\label{eq:overview}
\end{equation}
where $\hat{Y}^{s} \in \mathbb{R}^{N \times H}$ is the base forecast, $\hat{\varepsilon}$ the estimated innovations, $\Delta \in \mathbb{R}^{N \times H}$ the correction, $\alpha \in [0, 1]^{N}$ the target-wise combination weight, and $\odot$ row-wise multiplication.

\subsection{Shared-Component Pathway (SCP)}
\label{sec:method-scp}

The Shared-Component Pathway (SCP, Fig.~\ref{fig:method}(a)) estimates the common component holistically over all variables, and its structure mirrors the decomposition $c_j = p_j + \lambda_j^{\top} f$ in Definition~\ref{def:common}: the periodic basis table estimates the periodic part $p_j$, and cross-sectional pooling with common factors estimates the factor part $\lambda_j^{\top} f$; all components are shared by all targets and do not vary with the target. Let $Z \in \mathbb{R}^{N \times L}$ be the input after window-wise normalization and $\phi(t)$ the phase of time $t$ within a period of length $T_p$.

\textbf{Periodic basis table.} The periodic part $p_j$ is represented by a learnable basis for each variable and each phase, which is added to the normalized input:
\begin{equation}
\tilde{Z}_j = Z_j + \mathcal{T}_j\bigl[\phi(t-L+1{:}t)\bigr],
\label{eq:pbt}
\end{equation}
where $\mathcal{T} \in \mathbb{R}^{N \times T_p}$ is the periodic basis table. The seasonal term $\mathcal{S}_j$ is a linear combination of the $M$ most recent same-phase values before the forecast origin and represents the amplitude of the periodic component in the current cycle.

\textbf{Cross-sectional pooling and common factors.} The factor part $\lambda_j^{\top} f$ is represented jointly by the cross-sectional mean and $K$ common factors: the mean corresponds to the principal direction shared by all variables, and the factors to the remaining common directions with different proportions:
\begin{equation}
m^{c}_i = W_o\Bigl[\frac{1}{N}\sum_{j=1}^{N} W_v Z_j + \sum_{k=1}^{K} g_k \Lambda_{ik} \sum_{j=1}^{N} \Omega_{kj} W_v Z_j\Bigr],
\label{eq:pool}
\end{equation}
where $W_v$ and $W_o$ are learnable projections; $\Omega \in \mathbb{R}^{K \times N}$ and $\Lambda \in \mathbb{R}^{N \times K}$ are the pooling weights and loadings of the factors; and $g_k$ is a per-factor gate. The pooling weights and loadings are static parameters shared by all targets.

\textbf{Base forecast.} The base forecast combines the representations of the common component with a linear autoregressive term for the target itself:
\begin{equation}
\begin{aligned}
U_j &= \mathrm{Proj}\bigl(\tilde{Z}_j + m^{c}_j\bigr), \\
\hat{Y}^{s}_j &= \mathrm{Denorm}\bigl(\mathrm{Head}(\mathrm{MLP}(U_j) + U_j) \\
&\qquad + W_{\mathrm{lin}} Z_j + \mathcal{S}_j\bigr),
\end{aligned}
\label{eq:base}
\end{equation}
where $\mathrm{Proj}: \mathbb{R}^{L} \to \mathbb{R}^{d}$ and $\mathrm{Head}: \mathbb{R}^{d} \to \mathbb{R}^{H}$ are linear maps shared by all variables, $W_{\mathrm{lin}} \in \mathbb{R}^{H \times L}$ is the linear autoregressive bypass, and $\mathrm{Denorm}$ inverts the window-wise normalization.

\subsection{Innovation Pathway (ISP)}
\label{sec:method-isp}

The Innovation-Space Pathway (ISP, Fig.~\ref{fig:method}(b)--(d)) reads dependencies for each target in the innovation space, and its structure mirrors Definitions~\ref{def:common} and~\ref{def:dep}: prewhitening obtains innovations from the observations as in Definition~\ref{def:common}, and source--lag routing with the transfer readout selects sources and reads dependencies on the innovations as in Definition~\ref{def:dep}. We describe it for target $i$.

\textbf{Prewhitening.} Following Definition~\ref{def:common}, prewhitening proceeds in two steps: the common component is first removed with a periodic profile and cross-sectional factors, and the own dynamics of the idiosyncratic component are then removed with a $P$-th order autoregressive filter:
\begin{equation}
\begin{aligned}
r_j(t) &= x_j(t) - \psi_j[\phi(t)], \qquad e_j(t) = r_j(t) - \hat{\lambda}_j^{\top}\hat{f}(t), \\
\hat{\varepsilon}_j(t) &= \frac{1}{\sigma_j}\Bigl[e_j(t) - \sum_{p=1}^{P}\Phi_{jp}\, e_j(t-p)\Bigr],
\end{aligned}
\label{eq:prewhiten}
\end{equation}
where $\psi \in \mathbb{R}^{N \times T_p}$ is the periodic profile, $\hat{f}(t) = \Omega^{\mathrm{pw}} r(t) \in \mathbb{R}^{K_{\mathrm{pw}}}$ are $K_{\mathrm{pw}}$ cross-sectional factors with loadings $\hat{\lambda}_j$, which reduce to the cross-sectional average when $K_{\mathrm{pw}} = 1$ with uniform pooling weights, $\Phi \in \mathbb{R}^{N \times P}$ are the autoregressive coefficients, and $\sigma_j$ is the standard deviation of the innovations on the training segment. The quantities $\psi$, $\hat{\lambda}$, and $\Phi$ are estimated on the training segment and refined by the training objective. The main results use $K_{\mathrm{pw}} = 1$, which requires no additional loadings to be estimated; multi-factor removal does not consistently reduce the error further (Section~\ref{sec:exp-ablation}). Since $\hat{f}$ only approximates $f$, $\hat{\varepsilon}_j$ is an estimate of the innovation $\varepsilon_j$ of Definition~\ref{def:common} and may retain part of the common component, which is quantified in Section~\ref{sec:exp-ablation}.

\textbf{Prior candidates.} The candidate sources and the source--lag prior are determined by the lagged correlations between training-segment innovations (Definition~\ref{def:dep}): for each target, the correlations $\hat{\rho}_{ik}(l)$ at lags $1$ to $\ell_{\max}$ are computed, $C$ candidate sources $\mathcal{C}_i$ are selected according to $\max_l |\hat{\rho}_{ik}(l)|$, and $|\hat{\rho}_{ik}(l)|$ serves as the source--lag prior.

\textbf{Spectral source retention.} A retention probability is assigned to each candidate in every window, based on a learnable distance between the target and the candidate on the normalized amplitude spectra of the innovations:
\begin{equation}
q_i = A\,\frac{|\mathcal{F}(\hat{\varepsilon}_i)|}{\lVert \mathcal{F}(\hat{\varepsilon}_i) \rVert}, \qquad p_{ik} = \sigma\bigl(\zeta - \lVert q_i - q_k \rVert^2\bigr), \qquad k \in \mathcal{C}_i,
\label{eq:retain}
\end{equation}
where $\mathcal{F}$ is the discrete Fourier transform, $A$ a learnable projection, and $\zeta$ a learnable threshold. During training, a retention mask $\mu_{ik}$ is drawn by relaxed Bernoulli sampling~\cite{maddison2017concrete,jang2017gumbel}; at inference, $\mu_{ik} = p_{ik}$.

\textbf{Source--lag routing.} The routing score consists of three terms, namely a learnable source--lag score table, the in-window correlation of innovations, and the retention mask:
\begin{equation}
\begin{aligned}
\mathrm{score}_{ikl} &= \Theta_{ikl} + \gamma\,\frac{\bigl\langle W_q \hat{\varepsilon}^{(S)}_i,\, W_k \hat{\varepsilon}^{(S)}_{k,l} \bigr\rangle}{S} + \log \mu_{ik}, \\
w_i &= \mathrm{Softmax}_{(k, l)}\bigl(\mathrm{score}_{i}\bigr),
\end{aligned}
\label{eq:route}
\end{equation}
where $\Theta$ is a learnable score table seeded with the source--lag prior and $\hat{\varepsilon}^{(S)}$ denotes innovations over the most recent $S$ steps; summing over lags gives the reading weights $P_{ik} = \sum_l w_{ikl}$.

\textbf{Rational transfer readout.} The readout represents the response of the target to source innovations by a rational transfer function with a first-order denominator: a geometric recursion is applied to the weighted sum of the selected source innovations, which is then mapped to each forecast step through a lag-shared transfer kernel:
\begin{equation}
\begin{aligned}
z_{il}(t) &= \sum_{k \in \mathcal{C}_i} w_{ikl}\, a_{ik}\, \hat{\varepsilon}_k(t), \qquad \tilde{z}_{il}(t) = \sum_{t' \le t} \delta_i^{\,t-t'} z_{il}(t'), \\
o_{ih} &= \sigma_i \sum_{l}\sum_{t} \tilde{z}_{il}(t)\, V_{lth},
\end{aligned}
\label{eq:transfer}
\end{equation}
where $a_{ik}$ is a pairwise reading coefficient, $\delta_i \in (-1, 1)$ the denominator coefficient, and $V$ the transfer kernel; the recursion corresponds to the denominator of the transfer function, and the kernel to its numerator and pure delay.

\textbf{Time- and horizon-aware modulation.} The correction is modulated for each forecast step and clipped, with modulation coefficients determined by the energy of the selected source and target innovations and by the phase of the forecast origin:
\begin{equation}
\begin{aligned}
\Delta_{ih} &= \beta_h\, \xi_{ih}\, 2\tanh(o_{ih}/2), \\
\xi_{ih} &= \sigma\bigl(\mathrm{MLP}_h(\log E^{\mathrm{sel}}_i, \log E^{\mathrm{self}}_i, \mathbf{c}_t)\bigr),
\end{aligned}
\label{eq:modulate}
\end{equation}
where $\xi_{ih}$ is the modulation gate, $E^{\mathrm{sel}}_i$ and $E^{\mathrm{self}}_i$ are the energies of the selected source innovations and of the target's own innovations, $\mathbf{c}_t$ is the daily and weekly phase of the forecast origin, and $\beta_h$ is a step-wise coefficient; the correction acts on the first $\min(\ell_{\max}, H)$ forecast steps.

\subsection{Combination Weight}
\label{sec:method-alpha}

After training, the combination weight of each target is solved in closed form on the validation set:
\begin{equation}
\alpha_i = \mathrm{clip}_{[0, 1]}\Bigl(\frac{\sum_{\mathrm{val}} \langle Y_i - \hat{Y}^{s}_i,\, \Delta_i \rangle}{\sum_{\mathrm{val}} \lVert \Delta_i \rVert^2}\Bigr),
\label{eq:alpha}
\end{equation}
which is the solution of $\min_{\alpha} \sum_{\mathrm{val}} \lVert Y_i - \hat{Y}^{s}_i - \alpha \Delta_i \rVert^2$ over $[0, 1]$.

\subsection{Training Objective}
\label{sec:method-loss}

The training objective of PRELUDE is
\begin{equation}
\mathcal{L} = \ell\bigl(Y, \hat{Y}^{s} + \Delta\bigr) + \lambda_{\mathrm{1s}}\, \overline{\hat{\varepsilon}^{\,2}} + \lambda_{\mathrm{ret}}\, \frac{1}{N}\sum_{i}\Bigl(\frac{1}{C}\sum_{k \in \mathcal{C}_i} p_{ik} - r\Bigr)^2,
\label{eq:loss}
\end{equation}
where the first term is the forecasting loss with an element-wise regression loss $\ell$; the second term is the one-step prediction loss of the innovations, which drives prewhitening toward the residual of the best linear predictor; and the third term is a retention-ratio constraint that keeps, on average, about a fraction $r$ of the candidate sources for each target. The shared-component pathway is trained first and then frozen, after which the innovation pathway is trained; hyperparameter values are given in Section~\ref{sec:exp-setup}.

\section{Experiments}
\label{sec:exp}

\subsection{Experimental Setup}
\label{sec:exp-setup}

\textbf{Datasets.} We conduct experiments on two groups of settings (Table~\ref{tab:datasets}): the benchmark setting contains 12 widely used datasets from traffic, energy, and weather~\cite{song2020stsgcn,lai2018lstnet,zhou2021informer,wu2021autoformer}; the short-horizon and air-quality setting contains short-horizon forecasting on the traffic data and four air-quality datasets~\cite{wang2020pm25gnn,wang2025knowairv2,godahewa2021monash}. The PEMS series, the ETT series, and the air-quality data are split chronologically at 6:2:2 (12/4/4 months for ETT), and the other datasets at 7:1:2.

\begin{table}[!t]
\caption{Dataset Statistics. $N$ is the number of variables, $T$ the series length, and $T_p$ the period length.}
\label{tab:datasets}
\centering
\scriptsize
\setlength{\tabcolsep}{2.2pt}
\resizebox{\columnwidth}{!}{%
\begin{tabular}{llrrlrl}
\toprule
Dataset & Domain & $N$ & $T$ & Interval & $T_p$ & Horizons \\
\midrule
\multicolumn{7}{l}{\emph{Benchmark setting}} \\
PEMS03/04/07/08 & Traffic & 170--883 & 16,992--28,224 & 5 min & 288 & 12--96 \\
Traffic & Traffic & 862 & 17,544 & 1 h & 24 & 96--720 \\
Electricity & Energy & 321 & 26,304 & 1 h & 24 & 96--720 \\
Solar & Energy & 137 & 52,560 & 10 min & 144 & 96--720 \\
ETTh1/h2/m1/m2 & Energy & 7 & 17,420--69,680 & 1 h/15 min & 24/96 & 96--720 \\
Weather & Weather & 21 & 52,696 & 10 min & 144 & 96--720 \\
\midrule
\multicolumn{7}{l}{\emph{Short-horizon and air-quality setting}} \\
PEMS03/04/07/08 & Traffic & as above & as above & 5 min & 288 & 3, 6 \\
KnowAir & Air quality & 184 & 11,688 & 3 h & 8 & 2--24 \\
BTHSA/YRD/KDD2018 & Air quality & 127--270 & 21,818--70,128 & 1 h & 24 & 3--24 \\
\bottomrule
\end{tabular}%
}
\end{table}

\textbf{Baselines.} We select a range of recent forecasting models as baselines and group them by the reading forms of Section~\ref{sec:related}: cross-sectional pooling, SOFTS~\cite{han2024softs} and TQNet~\cite{lin2025tqnet}; inter-variable attention, iTransformer~\cite{liu2024itransformer}, EMAformer~\cite{zhang2026emaformer}, FreEformer~\cite{yue2025freeformer}, DUET~\cite{qiu2025duet}, and Crisp~\cite{wang2026crisp}; graph structures, CGTFra~\cite{yu2026cgtfra} and TimeFilter~\cite{hu2025timefilter}; explicit lead--lag estimation, LIFT~\cite{zhao2024lift}; and channel-independent and other architectures, CycleNet~\cite{lin2024cyclenet}, TimeEmb~\cite{xia2025timeemb}, and PatchTST~\cite{nie2023patchtst}.

\textbf{Evaluation protocol.} Following prior work, we use MSE and MAE as evaluation metrics, with a look-back length of 96 throughout. Normalization statistics are estimated on the training segment only, and all windows of the test segment are evaluated without dropping the last batch. Baseline results are taken from the original papers under the same protocol, and unreported settings are obtained by running the official code under our protocol.

\textbf{Implementation details.} PRELUDE is implemented in PyTorch and trained with the Adam optimizer on NVIDIA L20 GPUs. The hyperparameters of the innovation pathway are fixed across all datasets: $P = 24$, $C = 16$, $\ell_{\max} = 72$, $S = 24$, and $r = 0.25$. The number of common factors $K \in \{0, 2, 4, 8\}$, the number of periods of the seasonal term $M \in \{0, 3\}$, and the remaining training hyperparameters are selected on the validation set only; the test set is not involved in any selection. Results of PRELUDE are averaged over three random seeds.

\subsection{Main Results}
\label{sec:exp-main}

Table~\ref{tab:main} reports the results on the benchmark setting averaged over the four horizons, and Table~\ref{tab:t2} reports the full results on the short-horizon and air-quality setting. We make the following observations.

\textbf{1) PRELUDE achieves the best overall performance.} Across the 12 datasets of the benchmark setting, PRELUDE achieves the lowest average MSE on 8 and the lowest average MAE on 7, with average ranks of 1.75 and 1.58 for MSE and MAE. Among the 24 settings of the short-horizon and air-quality setting, PRELUDE achieves the best MSE on 23 and the best MAE on 20, with average ranks of 1.08 and 1.25, and reduces the average MSE by 6.1\% relative to the strongest baseline. One-sided Wilcoxon signed-rank tests, paired over the 48 dataset--horizon combinations of Table~\ref{tab:main} and the 24 settings of Table~\ref{tab:t2} with ties discarded, show that the improvements of PRELUDE in MSE and MAE over every baseline are significant ($p < 10^{-4}$).

\textbf{2) PRELUDE yields consistent improvements over baselines of every reading form.} Relative to SOFTS, TQNet, iTransformer, EMAformer, DUET, and CGTFra, PRELUDE reduces MSE on the benchmark setting by 5.4\%--16.5\% on average; relative to LIFT, which explicitly estimates lead--lag relations, it reduces MSE by 15.6\% on the short-horizon and air-quality setting.

\begin{table*}[t]
\caption{Multivariate Forecasting Results on the Benchmark Setting (Look-Back 96, Averaged over the Four Horizons). The best results are in bold and the second best are underlined; LIFT diverges on ETTh2, ETTm1, and ETTm2 at long horizons. Per-horizon results are provided in the code repository.}
\label{tab:main}
\centering
\resizebox{\textwidth}{!}{%
\begin{tabular}{llcccccccccccccccccccccccc}
\toprule
\multicolumn{2}{c}{Model} & \multicolumn{2}{c}{PRELUDE} & \multicolumn{2}{c}{TQNet} & \multicolumn{2}{c}{EMAformer} & \multicolumn{2}{c}{CGTFra} & \multicolumn{2}{c}{Crisp} & \multicolumn{2}{c}{TimeEmb} & \multicolumn{2}{c}{DUET} & \multicolumn{2}{c}{CycleNet} & \multicolumn{2}{c}{SOFTS} & \multicolumn{2}{c}{iTransformer} & \multicolumn{2}{c}{PatchTST} & \multicolumn{2}{c}{LIFT} \\
\cmidrule(lr){3-4}\cmidrule(lr){5-6}\cmidrule(lr){7-8}\cmidrule(lr){9-10}\cmidrule(lr){11-12}\cmidrule(lr){13-14}\cmidrule(lr){15-16}\cmidrule(lr){17-18}\cmidrule(lr){19-20}\cmidrule(lr){21-22}\cmidrule(lr){23-24}\cmidrule(lr){25-26}
\multicolumn{2}{c}{Metric} & MSE & MAE & MSE & MAE & MSE & MAE & MSE & MAE & MSE & MAE & MSE & MAE & MSE & MAE & MSE & MAE & MSE & MAE & MSE & MAE & MSE & MAE & MSE & MAE \\
\midrule
PEMS03 & Avg & \textbf{0.086} & \textbf{0.183} & 0.097 & 0.203 & \underline{0.090} & \underline{0.190} & 0.108 & 0.212 & 0.113 & 0.214 & 0.142 & 0.239 & 0.113 & 0.219 & 0.118 & 0.225 & 0.104 & 0.210 & 0.113 & 0.222 & 0.180 & 0.291 & 0.138 & 0.246 \\
\midrule
PEMS04 & Avg & \textbf{0.078} & \textbf{0.174} & 0.091 & 0.196 & \underline{0.081} & \underline{0.180} & 0.105 & 0.209 & 0.111 & 0.217 & 0.120 & 0.224 & 0.109 & 0.220 & 0.119 & 0.232 & 0.102 & 0.208 & 0.111 & 0.221 & 0.195 & 0.307 & 0.145 & 0.256 \\
\midrule
PEMS07 & Avg & \textbf{0.066} & \textbf{0.151} & 0.074 & 0.171 & \underline{0.073} & \underline{0.162} & 0.094 & 0.188 & 0.091 & 0.185 & 0.104 & 0.196 & 0.095 & 0.197 & 0.113 & 0.214 & 0.086 & 0.184 & 0.101 & 0.204 & 0.211 & 0.303 & 0.146 & 0.244 \\
\midrule
PEMS08 & Avg & \textbf{0.103} & \textbf{0.189} & 0.142 & 0.230 & 0.128 & \underline{0.209} & 0.145 & 0.225 & 0.152 & 0.229 & 0.154 & 0.229 & \underline{0.123} & 0.212 & 0.150 & 0.246 & 0.138 & 0.220 & 0.150 & 0.226 & 0.280 & 0.321 & 0.186 & 0.267 \\
\midrule
Traffic & Avg & 0.433 & 0.260 & 0.445 & 0.276 & 0.430 & \textbf{0.255} & \underline{0.427} & \underline{0.257} & 0.482 & 0.280 & 0.454 & 0.293 & 0.451 & 0.269 & 0.472 & 0.301 & \textbf{0.409} & 0.267 & 0.428 & 0.282 & 0.481 & 0.300 & 0.562 & 0.357 \\
\midrule
Electricity & Avg & \underline{0.161} & \underline{0.249} & 0.165 & 0.259 & \textbf{0.158} & \textbf{0.247} & 0.165 & 0.253 & \underline{0.161} & 0.253 & 0.167 & 0.260 & 0.172 & 0.259 & 0.168 & 0.259 & 0.174 & 0.264 & 0.178 & 0.270 & 0.205 & 0.290 & 0.221 & 0.309 \\
\midrule
Solar & Avg & \textbf{0.186} & \underline{0.226} & 0.198 & 0.256 & \underline{0.197} & \textbf{0.224} & 0.224 & 0.228 & 0.222 & 0.229 & 0.254 & 0.266 & 0.237 & 0.233 & 0.210 & 0.261 & 0.229 & 0.256 & 0.233 & 0.262 & 0.270 & 0.307 & 0.292 & 0.322 \\
\midrule
ETTh1 & Avg & 0.433 & 0.426 & 0.441 & 0.434 & \underline{0.432} & \textbf{0.424} & 0.434 & 0.427 & 0.436 & 0.430 & \textbf{0.425} & \underline{0.425} & 0.443 & 0.436 & 0.457 & 0.441 & 0.449 & 0.443 & 0.454 & 0.448 & 0.469 & 0.455 & 0.462 & 0.450 \\
\midrule
ETTh2 & Avg & 0.366 & \underline{0.391} & 0.378 & 0.402 & 0.374 & 0.395 & 0.369 & 0.394 & \textbf{0.361} & 0.392 & \underline{0.362} & \textbf{0.390} & 0.372 & 0.397 & 0.388 & 0.409 & 0.373 & 0.400 & 0.383 & 0.407 & 0.387 & 0.407 & 343.2 & 1.592 \\
\midrule
ETTm1 & Avg & \textbf{0.348} & \textbf{0.366} & 0.377 & 0.393 & 0.381 & \underline{0.380} & 0.388 & 0.387 & 0.373 & 0.381 & \underline{0.368} & 0.384 & 0.390 & 0.393 & 0.379 & 0.396 & 0.393 & 0.402 & 0.407 & 0.410 & 0.387 & 0.400 & 3.785 & 0.623 \\
\midrule
ETTm2 & Avg & \textbf{0.254} & \textbf{0.304} & 0.277 & 0.323 & 0.271 & 0.313 & 0.277 & 0.317 & 0.268 & 0.313 & \underline{0.265} & \underline{0.308} & 0.280 & 0.324 & 0.266 & 0.314 & 0.287 & 0.330 & 0.288 & 0.332 & 0.281 & 0.326 & 1421.0 & 1.323 \\
\midrule
Weather & Avg & \textbf{0.221} & \textbf{0.252} & 0.242 & 0.269 & 0.240 & 0.262 & 0.237 & \underline{0.261} & \underline{0.235} & \underline{0.261} & 0.237 & 0.262 & 0.251 & 0.273 & 0.243 & 0.271 & 0.255 & 0.278 & 0.258 & 0.278 & 0.259 & 0.273 & 0.260 & 0.300 \\
\bottomrule
\end{tabular}
}
\vspace{4pt}
\caption{Multivariate Forecasting Results on the Short-Horizon and Air-Quality Setting (Look-Back 96). The best results are in bold and the second best are underlined.}
\label{tab:t2}
\centering
\resizebox{\textwidth}{!}{%
\begin{tabular}{llcccccccccccccccccccccccc}
\toprule
\multicolumn{2}{c}{Model} & \multicolumn{2}{c}{PRELUDE} & \multicolumn{2}{c}{TQNet} & \multicolumn{2}{c}{EMAformer} & \multicolumn{2}{c}{CGTFra} & \multicolumn{2}{c}{TimeFilter} & \multicolumn{2}{c}{FreEformer} & \multicolumn{2}{c}{DUET} & \multicolumn{2}{c}{CycleNet} & \multicolumn{2}{c}{SOFTS} & \multicolumn{2}{c}{iTransformer} & \multicolumn{2}{c}{PatchTST} & \multicolumn{2}{c}{LIFT} \\
\cmidrule(lr){3-4}\cmidrule(lr){5-6}\cmidrule(lr){7-8}\cmidrule(lr){9-10}\cmidrule(lr){11-12}\cmidrule(lr){13-14}\cmidrule(lr){15-16}\cmidrule(lr){17-18}\cmidrule(lr){19-20}\cmidrule(lr){21-22}\cmidrule(lr){23-24}\cmidrule(lr){25-26}
\multicolumn{2}{c}{Metric} & MSE & MAE & MSE & MAE & MSE & MAE & MSE & MAE & MSE & MAE & MSE & MAE & MSE & MAE & MSE & MAE & MSE & MAE & MSE & MAE & MSE & MAE & MSE & MAE \\
\midrule
KnowAir & 2 & \textbf{0.1250} & \textbf{0.2064} & 0.1394 & 0.2212 & 0.1341 & 0.2140 & 0.1294 & \underline{0.2096} & \underline{0.1280} & 0.2117 & 0.1365 & 0.2183 & 0.1386 & 0.2182 & 0.1438 & 0.2274 & 0.1380 & 0.2189 & 0.1366 & 0.2183 & 0.1443 & 0.2261 & 0.1404 & 0.2248 \\
 & 4 & \textbf{0.1684} & \textbf{0.2475} & 0.1872 & 0.2646 & 0.1803 & 0.2558 & \underline{0.1752} & \underline{0.2515} & 0.1756 & 0.2549 & 0.1889 & 0.2657 & 0.1832 & 0.2582 & 0.1933 & 0.2718 & 0.1855 & 0.2635 & 0.1817 & 0.2598 & 0.1950 & 0.2747 & 0.1891 & 0.2709 \\
 & 8 & \textbf{0.2209} & \textbf{0.2912} & 0.2430 & 0.3099 & 0.2351 & 0.2995 & \underline{0.2283} & \underline{0.2953} & 0.2436 & 0.3075 & 0.2351 & 0.2993 & 0.2316 & 0.2979 & 0.2486 & 0.3181 & 0.2393 & 0.3078 & 0.2333 & 0.3021 & 0.2494 & 0.3207 & 0.2454 & 0.3154 \\
 & 24 & \textbf{0.3166} & \textbf{0.3633} & 0.3380 & 0.3765 & 0.3436 & 0.3708 & 0.3410 & 0.3696 & 0.3578 & 0.3882 & 0.3386 & \underline{0.3660} & 0.3315 & 0.3669 & 0.3342 & 0.3819 & 0.3372 & 0.3754 & \underline{0.3262} & 0.3695 & 0.3503 & 0.4011 & 0.3368 & 0.3883 \\
 & Avg & \textbf{0.2078} & \textbf{0.2771} & 0.2269 & 0.2931 & 0.2233 & 0.2850 & \underline{0.2185} & \underline{0.2815} & 0.2262 & 0.2905 & 0.2248 & 0.2873 & 0.2212 & 0.2853 & 0.2300 & 0.2998 & 0.2250 & 0.2914 & 0.2195 & 0.2874 & 0.2348 & 0.3057 & 0.2279 & 0.2998 \\
\midrule
BTHSA & 3 & \textbf{0.0522} & 0.1282 & 0.0590 & 0.1371 & 0.0538 & \underline{0.1281} & \underline{0.0523} & \textbf{0.1269} & 0.0555 & 0.1328 & 0.0584 & 0.1386 & 0.0545 & 0.1289 & 0.0641 & 0.1430 & 0.0596 & 0.1387 & 0.0549 & 0.1314 & 0.0671 & 0.1474 & 0.0644 & 0.1427 \\
 & 6 & \textbf{0.0869} & \textbf{0.1670} & 0.0997 & 0.1792 & 0.0912 & \underline{0.1676} & \underline{0.0903} & \underline{0.1676} & 0.0941 & 0.1743 & 0.0931 & 0.1756 & 0.0942 & 0.1700 & 0.1097 & 0.1894 & 0.1027 & 0.1827 & 0.0951 & 0.1742 & 0.1140 & 0.1942 & 0.1096 & 0.1905 \\
 & 12 & \textbf{0.1442} & \underline{0.2195} & 0.1620 & 0.2337 & \underline{0.1491} & \textbf{0.2188} & 0.1520 & 0.2226 & 0.1705 & 0.2386 & 0.1549 & 0.2272 & 0.1515 & 0.2231 & 0.1789 & 0.2508 & 0.1644 & 0.2363 & 0.1612 & 0.2358 & 0.1833 & 0.2600 & 0.1780 & 0.2515 \\
 & 24 & \textbf{0.2245} & \underline{0.2813} & 0.2443 & 0.2949 & \underline{0.2335} & \textbf{0.2802} & 0.2402 & 0.2877 & 0.2808 & 0.3135 & 0.2422 & 0.2901 & 0.2387 & 0.2862 & 0.2622 & 0.3216 & 0.2520 & 0.2997 & 0.2467 & 0.3001 & 0.2662 & 0.3191 & 0.2612 & 0.3205 \\
 & Avg & \textbf{0.1269} & \underline{0.1990} & 0.1412 & 0.2112 & \underline{0.1319} & \textbf{0.1987} & 0.1337 & 0.2012 & 0.1502 & 0.2148 & 0.1371 & 0.2079 & 0.1347 & 0.2020 & 0.1538 & 0.2262 & 0.1447 & 0.2144 & 0.1395 & 0.2104 & 0.1577 & 0.2302 & 0.1533 & 0.2263 \\
\midrule
YRD & 3 & 0.0616 & 0.1538 & 0.0693 & 0.1639 & 0.0632 & 0.1542 & \underline{0.0614} & \textbf{0.1535} & \textbf{0.0610} & \underline{0.1536} & 0.0669 & 0.1621 & 0.0641 & 0.1557 & 0.0761 & 0.1730 & 0.0697 & 0.1656 & 0.0651 & 0.1583 & 0.0767 & 0.1724 & 0.0754 & 0.1712 \\
 & 6 & \textbf{0.1015} & \textbf{0.1999} & 0.1147 & 0.2134 & 0.1055 & \underline{0.2015} & \underline{0.1041} & 0.2020 & 0.1102 & 0.2100 & 0.1079 & 0.2094 & 0.1066 & 0.2040 & 0.1282 & 0.2280 & 0.1156 & 0.2151 & 0.1111 & 0.2098 & 0.1298 & 0.2313 & 0.1252 & 0.2257 \\
 & 12 & \textbf{0.1623} & \textbf{0.2595} & 0.1786 & 0.2762 & 0.1697 & \underline{0.2617} & 0.1738 & 0.2663 & 0.1870 & 0.2796 & \underline{0.1686} & 0.2643 & 0.1717 & 0.2654 & 0.1959 & 0.2938 & 0.1815 & 0.2775 & 0.1792 & 0.2750 & 0.2004 & 0.3006 & 0.1937 & 0.2905 \\
 & 24 & \textbf{0.2337} & \textbf{0.3201} & 0.2517 & 0.3322 & 0.2519 & 0.3261 & 0.2502 & 0.3280 & 0.2694 & 0.3435 & \underline{0.2448} & 0.3245 & 0.2479 & \underline{0.3241} & 0.2666 & 0.3536 & 0.2543 & 0.3347 & 0.2542 & 0.3337 & 0.2728 & 0.3615 & 0.2651 & 0.3526 \\
 & Avg & \textbf{0.1398} & \textbf{0.2333} & 0.1536 & 0.2465 & 0.1476 & \underline{0.2359} & 0.1474 & 0.2375 & 0.1569 & 0.2467 & \underline{0.1470} & 0.2401 & 0.1476 & 0.2373 & 0.1667 & 0.2621 & 0.1553 & 0.2482 & 0.1524 & 0.2442 & 0.1699 & 0.2664 & 0.1648 & 0.2600 \\
\midrule
KDD2018 & 3 & \textbf{0.0943} & \textbf{0.1285} & 0.1112 & 0.1461 & 0.1009 & 0.1369 & \underline{0.1000} & \underline{0.1324} & 0.1081 & 0.1495 & 0.1114 & 0.1525 & 0.1061 & 0.1393 & 0.1123 & 0.1521 & 0.1158 & 0.1523 & 0.1080 & 0.1491 & 0.1203 & 0.1526 & 0.1147 & 0.1517 \\
 & 6 & \textbf{0.1645} & \textbf{0.1847} & 0.1960 & 0.2107 & \underline{0.1774} & 0.1927 & 0.1782 & \underline{0.1925} & 0.1989 & 0.2164 & 0.1870 & 0.2047 & 0.1896 & 0.1952 & 0.1915 & 0.2134 & 0.1921 & 0.2120 & 0.1901 & 0.2085 & 0.2015 & 0.2152 & 0.1995 & 0.2143 \\
 & 12 & \textbf{0.2929} & \textbf{0.2639} & 0.3199 & 0.2871 & \underline{0.3073} & \underline{0.2714} & 0.3218 & 0.2780 & 0.3561 & 0.3068 & 0.3162 & 0.2813 & 0.3240 & 0.2760 & 0.3179 & 0.2933 & 0.3281 & 0.2958 & 0.3250 & 0.2933 & 0.3374 & 0.2993 & 0.3275 & 0.2970 \\
 & 24 & \textbf{0.4556} & \textbf{0.3585} & 0.4880 & 0.3797 & 0.4788 & \underline{0.3658} & 0.4808 & 0.3693 & 0.5582 & 0.4065 & \underline{0.4778} & 0.3690 & 0.4922 & 0.3692 & 0.4804 & 0.3837 & 0.5022 & 0.3931 & 0.5104 & 0.3960 & 0.5153 & 0.3974 & 0.4890 & 0.3833 \\
 & Avg & \textbf{0.2518} & \textbf{0.2339} & 0.2788 & 0.2559 & \underline{0.2661} & \underline{0.2417} & 0.2702 & 0.2430 & 0.3053 & 0.2698 & 0.2731 & 0.2519 & 0.2780 & 0.2449 & 0.2756 & 0.2606 & 0.2845 & 0.2633 & 0.2834 & 0.2617 & 0.2937 & 0.2661 & 0.2827 & 0.2616 \\
\midrule
PEMS03 & 3 & \textbf{0.0404} & \textbf{0.1310} & 0.0433 & 0.1386 & \underline{0.0426} & \underline{0.1378} & \underline{0.0426} & 0.1383 & 0.0441 & 0.1405 & 0.0511 & 0.1548 & 0.0456 & 0.1427 & 0.0461 & 0.1459 & 0.0470 & 0.1467 & 0.0487 & 0.1488 & 0.0468 & 0.1464 & 0.0497 & 0.1511 \\
 & 6 & \textbf{0.0470} & \textbf{0.1396} & 0.0500 & 0.1475 & \underline{0.0492} & \underline{0.1464} & 0.0497 & 0.1471 & 0.0508 & 0.1492 & 0.0611 & 0.1734 & 0.0530 & 0.1524 & 0.0536 & 0.1559 & 0.0557 & 0.1579 & 0.0595 & 0.1629 & 0.0561 & 0.1586 & 0.0586 & 0.1625 \\
 & Avg & \textbf{0.0437} & \textbf{0.1353} & 0.0467 & 0.1430 & \underline{0.0459} & \underline{0.1421} & 0.0462 & 0.1427 & 0.0474 & 0.1449 & 0.0561 & 0.1641 & 0.0493 & 0.1475 & 0.0498 & 0.1509 & 0.0514 & 0.1523 & 0.0541 & 0.1559 & 0.0515 & 0.1525 & 0.0542 & 0.1568 \\
\midrule
PEMS04 & 3 & \textbf{0.0540} & \textbf{0.1458} & \underline{0.0567} & 0.1531 & 0.0586 & 0.1542 & 0.0577 & \underline{0.1530} & 0.0582 & 0.1547 & 0.0680 & 0.1812 & 0.0609 & 0.1585 & 0.0603 & 0.1596 & 0.0603 & 0.1575 & 0.0650 & 0.1658 & 0.0622 & 0.1629 & 0.0628 & 0.1638 \\
 & 6 & \textbf{0.0584} & \textbf{0.1511} & \underline{0.0614} & \underline{0.1595} & 0.0646 & 0.1620 & 0.0632 & 0.1604 & 0.0635 & 0.1617 & 0.0830 & 0.2022 & 0.0693 & 0.1697 & 0.0667 & 0.1692 & 0.0682 & 0.1678 & 0.0754 & 0.1792 & 0.0712 & 0.1750 & 0.0705 & 0.1744 \\
 & Avg & \textbf{0.0562} & \textbf{0.1485} & \underline{0.0591} & \underline{0.1563} & 0.0616 & 0.1581 & 0.0605 & 0.1567 & 0.0608 & 0.1582 & 0.0755 & 0.1917 & 0.0651 & 0.1641 & 0.0635 & 0.1644 & 0.0643 & 0.1627 & 0.0702 & 0.1725 & 0.0667 & 0.1690 & 0.0667 & 0.1691 \\
\midrule
PEMS07 & 3 & \textbf{0.0370} & \textbf{0.1185} & 0.0405 & 0.1271 & 0.0418 & 0.1286 & 0.0408 & \underline{0.1266} & \underline{0.0399} & \underline{0.1266} & 0.0505 & 0.1572 & 0.0450 & 0.1350 & 0.0432 & 0.1345 & 0.0418 & 0.1292 & 0.0462 & 0.1392 & 0.0446 & 0.1365 & 0.0458 & 0.1398 \\
 & 6 & \textbf{0.0418} & \textbf{0.1254} & 0.0460 & 0.1347 & 0.0477 & 0.1370 & 0.0468 & 0.1352 & \underline{0.0454} & \underline{0.1345} & 0.0571 & 0.1578 & 0.0511 & 0.1440 & 0.0501 & 0.1450 & 0.0490 & 0.1393 & 0.0555 & 0.1532 & 0.0533 & 0.1494 & 0.0539 & 0.1515 \\
 & Avg & \textbf{0.0394} & \textbf{0.1219} & 0.0433 & 0.1309 & 0.0448 & 0.1328 & 0.0438 & 0.1309 & \underline{0.0426} & \underline{0.1305} & 0.0538 & 0.1575 & 0.0481 & 0.1395 & 0.0466 & 0.1397 & 0.0454 & 0.1342 & 0.0509 & 0.1462 & 0.0489 & 0.1429 & 0.0499 & 0.1456 \\
\midrule
PEMS08 & 3 & \textbf{0.0489} & \textbf{0.1365} & 0.0523 & 0.1473 & 0.0519 & 0.1454 & \underline{0.0512} & \underline{0.1439} & 0.0522 & 0.1456 & 0.0619 & 0.1730 & 0.0540 & 0.1495 & 0.0563 & 0.1554 & 0.0526 & 0.1482 & 0.0568 & 0.1557 & 0.0567 & 0.1562 & 0.0573 & 0.1574 \\
 & 6 & \textbf{0.0554} & \textbf{0.1451} & 0.0585 & 0.1555 & 0.0589 & 0.1542 & \underline{0.0576} & \underline{0.1525} & 0.0586 & 0.1530 & 0.0786 & 0.1897 & 0.0627 & 0.1600 & 0.0649 & 0.1660 & 0.0599 & 0.1557 & 0.0677 & 0.1694 & 0.0683 & 0.1704 & 0.0673 & 0.1696 \\
 & Avg & \textbf{0.0521} & \textbf{0.1408} & 0.0554 & 0.1514 & 0.0554 & 0.1498 & \underline{0.0544} & \underline{0.1482} & 0.0554 & 0.1493 & 0.0702 & 0.1814 & 0.0584 & 0.1548 & 0.0606 & 0.1607 & 0.0562 & 0.1519 & 0.0623 & 0.1625 & 0.0625 & 0.1633 & 0.0623 & 0.1635 \\
\bottomrule
\end{tabular}
}
\end{table*}

\subsection{Sources of Gain}
\label{sec:exp-source}

Table~\ref{tab:source} decomposes the total error reduction of PRELUDE relative to a model with only the target's own dynamics into three shares: the periodic basis table and seasonal term, cross-sectional pooling and common factors, and the innovation pathway are added in turn, corresponding respectively to the periodic part $p_j$ of the common component, its factor part $\lambda_j^{\top} f$, and the source innovations of Definition~\ref{def:common}. The three shares separate clearly across data types: at least 87\% of the gain on rhythmic data comes from the periodic component, the common factors account for 87\%--93\% on air-quality data, and the dependency share is concentrated on short-horizon traffic (19\%--42\%). The correction of the innovation pathway acts only on the first $\min(\ell_{\max}, H)$ forecast steps and contributes nothing beyond them, so the gain on long horizons comes entirely from the shared-component pathway; within this range, the innovation pathway reduces the error by 0.4\%--3.9\% on short-horizon traffic, KnowAir, BTHSA, and YRD.

\begin{table}[!t]
\caption{Sources of Gain (Mean over Three Seeds). The three shares correspond to the periodic part and the factor part of the common component and to the source innovations of Definition~\ref{def:common}; the first two are obtained by the shared-component pathway and the third by the innovation pathway. Shares are signed contributions summing to 100\%, so a share may be negative or exceed 100\%.}
\label{tab:source}
\centering
\scriptsize
\renewcommand{\arraystretch}{0.9}
\setlength{\tabcolsep}{3pt}
\resizebox{\columnwidth}{!}{%
\begin{tabular}{lrrrrrr}
\toprule
Setting & Start & PRELUDE & Reduction & Periodic & Factor & Dependency \\
\midrule
ETTm1-96 & 0.3085 & 0.2791 & 9.5\% & 99\% & 1\% & 0\% \\
ETTm2-96 & 0.1705 & 0.1606 & 5.8\% & 103\% & $-$3\% & 0\% \\
Weather-96 & 0.1613 & 0.1444 & 10.5\% & 98\% & 2\% & 0\% \\
Electricity-96 & 0.1465 & 0.1298 & 11.4\% & 87\% & 13\% & 0\% \\
Traffic-96 & 0.4409 & 0.3976 & 9.8\% & 90\% & 11\% & 0\% \\
Solar-96 & 0.1990 & 0.1641 & 17.5\% & 68\% & 31\% & 0\% \\
PEMS08-12 & 0.0784 & 0.0647 & 17.5\% & 70\% & 27\% & 3\% \\
PEMS03-3 & 0.0436 & 0.0404 & 7.3\% & 26\% & 32\% & 42\% \\
PEMS04-3 & 0.0583 & 0.0540 & 7.4\% & 38\% & 43\% & 19\% \\
PEMS07-3 & 0.0410 & 0.0370 & 9.8\% & 35\% & 28\% & 37\% \\
PEMS08-3 & 0.0535 & 0.0489 & 8.6\% & 57\% & 20\% & 23\% \\
KnowAir-8 & 0.2469 & 0.2209 & 10.5\% & 7\% & 90\% & 4\% \\
BTHSA-6 & 0.1101 & 0.0869 & 21.1\% & 7\% & 87\% & 6\% \\
YRD-3 & 0.0775 & 0.0616 & 20.5\% & 7\% & 87\% & 6\% \\
KDD2018-3 & 0.1067 & 0.0943 & 11.6\% & 6\% & 93\% & 1\% \\
\bottomrule
\end{tabular}%
}
\end{table}

\begin{figure*}[t]
\centering
\includegraphics[width=\textwidth]{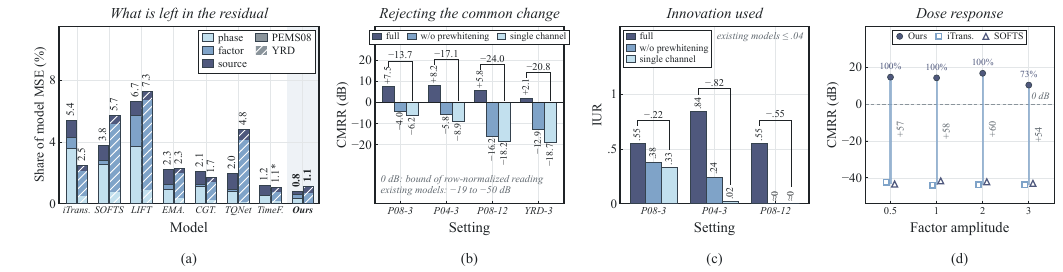}
\caption{Re-measurement of Section~\ref{sec:obs} on PRELUDE. (a) The part of the residual still predictable; (b) CMRR of the innovation pathway; (c) innovation utilization rate; (d) CMRR on synthetic data as the common factors strengthen. P08-3 denotes PEMS08 at horizon 3, and so on.}
\label{fig:remeasure}
\end{figure*}

\subsection{Ablation Study}
\label{sec:exp-ablation}

Table~\ref{tab:ablation} examines ten variants on six settings covering short-horizon traffic, air quality, medium-horizon traffic, and Electricity. The single-channel variant removes cross-sectional pooling and the common factors and applies the innovation pathway to the raw series, so that one set of row-normalized weights carries both kinds of information. We make the following observations.

\textbf{1) Handling the two kinds of information in different ways is necessary.} The single-channel variant shows the largest error increase on five of the six settings, reaching 15.7\% and 15.3\% on YRD and BTHSA. When both kinds of information are carried by one set of reading weights, the estimation--identification trade-off of Section~\ref{sec:theory-tradeoff} reappears, which PRELUDE avoids with two pathways.

\textbf{2) Each component carries its corresponding information.} Cross-sectional pooling and common factors are most effective on air-quality data, the periodic basis table on Electricity and medium-horizon traffic, and the innovation pathway and prior candidates act consistently on short-horizon traffic and air quality, in one-to-one correspondence with the gain composition in Table~\ref{tab:source}. A minimal innovation pathway with fixed prewhitening, prior candidates, and a linear readout performs on par with the full method on traffic data, and the learnable routing and readout reduce the error by a further 0.5\%--0.8\% on air-quality data: the gain comes mainly from separated reading in the innovation space rather than from stacking components.

\textbf{3) Prewhitening determines the reading property of the innovation pathway.} Without prewhitening, the CMRR of the innovation pathway turns from positive to negative and the read-in content is again dominated by the common component (Section~\ref{sec:exp-remeasure}); prewhitening is thus the prerequisite for the innovation pathway to read dependencies. Before removal, 61\%--88\% of the variance lies in the first eight common directions; removing the cross-sectional average lowers this to 10\%--22\% and raises the effective rank about tenfold, and removing several factors by PCA or learned factors lowers it further to 1\%--10\%, with an error on par with the cross-sectional average on traffic data. On synthetic data whose factor loadings are half positive and half negative, once the factors are strengthened to twice their amplitude or more, the hit rate of the planted sources drops to 12\%--54\% after removing the cross-sectional average, while PCA and learned-factor removal restore it to 100\%.

\begin{table}[!t]
\caption{Ablation Study (Mean over Three Seeds). The first row gives the MSE of PRELUDE and the other rows the relative change, where positive values indicate increased error; ``--'' indicates that the component is not used in that setting.}
\label{tab:ablation}
\centering
\scriptsize
\setlength{\tabcolsep}{2.4pt}
\resizebox{\columnwidth}{!}{%
\begin{tabular}{lrrrrrr}
\toprule
Variant & PEMS08-3 & PEMS04-3 & YRD-3 & BTHSA-3 & PEMS08-12 & Elec.-96 \\
\midrule
PRELUDE & 0.0489 & 0.0540 & 0.0616 & 0.0522 & 0.0647 & 0.1298 \\
\midrule
w/o periodic basis & +0.09\% & +0.36\% & $-$0.04\% & $-$0.79\% & +1.51\% & \textbf{+7.15\%} \\
w/o pooling \& factors & +1.27\% & +2.96\% & +11.98\% & +14.33\% & +4.90\% & +1.66\% \\
w/o common factors & -- & +0.62\% & +6.27\% & +7.83\% & $-$0.48\% & -- \\
w/o innovation pathway & +2.21\% & +1.50\% & +1.52\% & +2.02\% & +0.67\% & +0.00\% \\
w/o prewhitening & +0.47\% & +0.82\% & $-$0.19\% & +0.05\% & +0.22\% & $-$0.03\% \\
Random candidates & +1.85\% & +1.57\% & +0.56\% & +1.16\% & +0.52\% & +0.00\% \\
PCA removal & +0.04\% & +0.25\% & +1.01\% & $-$0.33\% & +0.01\% & $-$0.03\% \\
Learned-factor removal & 0.00\% & +0.03\% & +1.14\% & +0.74\% & $-$0.15\% & 0.00\% \\
Minimal innovation pathway & $-$0.40\% & +0.05\% & +0.45\% & +0.81\% & $-$0.19\% & +0.05\% \\
Single channel & \textbf{+2.24\%} & \textbf{+3.71\%} & \textbf{+15.71\%} & \textbf{+15.32\%} & \textbf{+5.09\%} & +1.40\% \\
\bottomrule
\end{tabular}%
}
\end{table}

\subsection{Mitigation of Spurious Dependency}
\label{sec:exp-remeasure}

Fig.~\ref{fig:remeasure} re-measures PRELUDE with the same diagnostics as Section~\ref{sec:obs}.

\textbf{Residuals of the two kinds of information.} The predictable part of the PRELUDE residual is 0.8\% on PEMS08 and 1.1\% on YRD, the lowest among all models (Fig.~\ref{fig:remeasure}(a)).

\textbf{Properties of reading.} The CMRR of the innovation pathway ranges from $+2.1$ to $+8.2$\,dB on the four settings, above the 0\,dB upper bound of row-normalized reading (Fig.~\ref{fig:remeasure}(b)); without prewhitening or with a single channel, it falls to between $-4.0$ and $-18.7$\,dB. When LIFT identifies lead--lag relations in the innovation space while still reading the raw series, the share of targets whose most-read source is the strongest source rises from 23\%--26\% to 33\%--47\% on traffic data, yet its CMRR stays around $-20$\,dB and its error increases by 4.1\%--12.6\%: the reading property is determined by the space in which reading takes place. On the three traffic settings, the IUR of PRELUDE is 0.55--0.84, against 0.00--0.07 with the shared-component pathway alone and at most 0.04 for existing models (Fig.~\ref{fig:remeasure}(c)); the first two differ only in the innovation pathway, so the IUR measures innovation use, not input sensitivity; as the common factors strengthen, the CMRR of the innovation pathway stays above $+10$\,dB (around $-43$\,dB for existing models, Fig.~\ref{fig:remeasure}(d)).

\textbf{Temporal alignment of the read content.} Circularly shifting the source innovations within the window at inference time breaks their alignment with the target; on six short-horizon traffic settings, the innovation pathway then changes from reducing the error by 1.3\%--4.1\% to increasing it by 0.3\%--2.9\%, so its gain comes from the lead--lag relation itself.

\subsection{Parameter Sensitivity}
\label{sec:exp-sens}

On the four air-quality datasets, the error decreases monotonically with the number of common factors $K$, and $K = 8$ reduces it by 3.8\%--6.9\% relative to $K = 0$; on short-horizon traffic, the error is insensitive to $K$. This corresponds to Proposition~\ref{prop:trunc}: when the common component is the superposition of several factors with different loadings, more common directions are needed.

\section{Conclusion}
\label{sec:conclusion}

This paper reveals spurious dependency in cross-variable forecasting: existing models gain from other variables mainly through the common component rather than through dependencies. We attribute it to common-component confounding and show, through interventions and theoretical analysis, that the common component and dependencies place different requirements on cross-variable reading and should therefore be handled in different ways. The resulting separated-reading framework, PRELUDE, estimates the common component holistically and reads dependencies for each target in the innovation space; on 16 real-world datasets from four domains, it effectively mitigates spurious dependency and achieves the best overall performance. Our analysis suggests that cross-variable modules should be evaluated by how much of each kind of information they obtain, not only by overall error.

\section*{Acknowledgment}
We used ChatGPT (OpenAI) and Claude (Anthropic) to assist in writing the plotting code for the data figures (Figs.~1--3 and 5); all figures are rendered from our experimental results, and the code was checked by the authors. We also used them for language editing. The authors take full responsibility for the correctness and originality of all content.

\bibliographystyle{IEEEtran}
\bibliography{references}

\end{document}